# Stochastic Learning of Nonstationary Kernels for Natural Language Modeling

**Sahil Garg** and **Greg Ver Steeg** and **Aram Galstyan**
USC Information Sciences Institute
Marina del Rey, CA 90292
{sahil,gregv,galstyan}@isi.edu

## Abstract

Natural language processing often involves computations with semantic or syntactic graphs to facilitate sophisticated reasoning based on structural relationships. While convolution kernels provide a powerful tool for comparing graph structure based on node (word) level relationships, they are difficult to customize and can be computationally expensive. We propose a generalization of convolution kernels, with a nonstationary model, for better expressibility of natural languages in supervised settings. For a scalable learning of the parameters introduced with our model, we propose a novel algorithm that leverages stochastic sampling on k-nearest neighbor graphs, along with approximations based on locality-sensitive hashing. We demonstrate the advantages of our approach on a challenging real-world (structured inference) problem of automatically extracting biological models from the text of scientific papers.

## 1 Introduction

Many problems in natural language processing (NLP) involve making *structured predictions* based on unstructured or semi-structured textual input. An important example of such a prediction problem is shown in Fig. 1. Here, a sentence is parsed into a semantic graph, from which we want to extract structured information on multiple biomolecular interactions, each involving interaction type, and two or three participants with a catalyst or domain role. The problem is formulated as binary classification of hypothesized candidates for multiple structured predictions (biomolecular interactions) from the semantic graph, corresponding to binary classification of discrete structures (acting as features for the problem), such as trees or paths. See Sec. 2 for more details on the problem. This NLP problem is highly relevant for building biological models on Cancerous biopathways (Cohen, 2015; Downward, 2003; Vogelstein and Kinzler, 2004).

An important family of classifiers that operate on discrete structures is based on convolution kernels, originally proposed by (Haussler, 1999), for computing similarity between discrete structures of any type and later extended for specific structures such as strings, sequences/paths, trees (Collins and Duffy, 2001; Zelenko et al., 2003; Mooney and Bunescu, 2005; Shi et al., 2009). Given a kernel similarity function and labeled data, a discrete structure can be classified using different methods such as kernel-k-NN (Yu et al., 2002), kernel-SVM (Cortes and Vapnik, 1995; Schölkopf and Smola, 2002), kernel-GP (Rasmussen, 2006) and so on.

Despite the success of convolution kernel-based methods in various NLP tasks (Collins and Duffy, 2002; Moschitti, 2006; Tikk et al., 2010; Qian and Zhou, 2012; Srivastava et al., 2013; Hovy et al., 2013; Filice et al., 2015; Tymoshenko et al., 2016), there are two important issues limiting their practicality in real-world applications. First, convolution kernels are not *flexible* enough to adequately model rich natural language representations, as they typically depend only on a few tunable parameters. This inherent rigidity prevents kernel-based methods from properly adapting to a given task, as opposed to, for instance, neural networks that typically have millions of parameters that are learned from data and show state-of-the-art results for a number of NLP problems (Collobert and Weston, 2008; Sundermeyer et al., 2012; Chen and Manning, 2014;

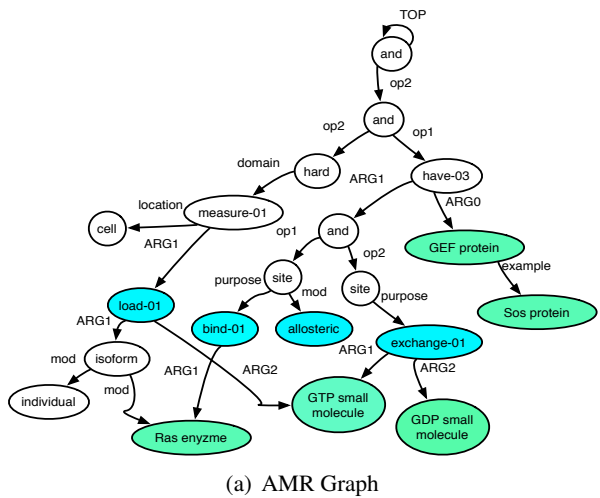

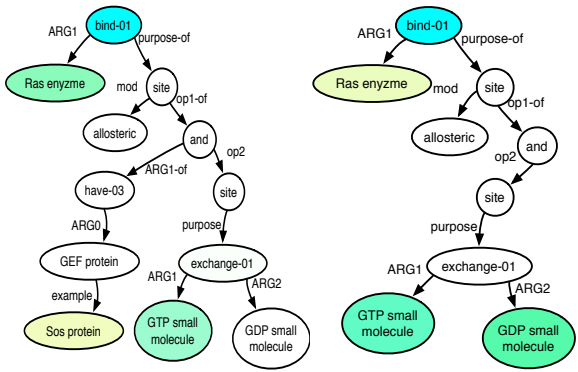

(b) correct (positive label)   (c) incorrect (negative label)

Figure 1: In 1(a), a sentence is parsed into a semantic graph. Then the two candidate hypothesis about different biomolecular interactions (structured prediction candidates) are generated automatically. According to the text, a valid hypothesis is that *Sos catalyzes binding between Ras and GTP*, while the alternative hypothesis *Ras catalyzes binding between GTP and GDP* is false; each of those hypotheses corresponds to one of the post-processed subgraphs shown in 1(b) and 1(c), respectively.

Sutskever et al., 2014; Kalchbrenner et al., 2014; Luong et al., 2015; Kumar et al., 2016). The second major issue with convolution kernels is the high computational cost, for both learning and inference, as computing kernel similarity between a pair of discrete structures costs polynomial time in its size. For instance, classifying a new data point based on $N$ labeled examples requires calculation of $N$ pairwise similarities, which might be prohibitively expensive for many real-world problems.

We address the first problem by proposing a nonstationary extension to the conventional convolution kernels, by introducing a novel, task-dependent parameterization of the kernel similarity function for better expressiveness and flexibility. Those parameters, which need to be inferred from the data, are defined in a way that allows the model to ignore substructures irrelevant for a given task when computing kernel similarity. For example, in Table 1, the screened-out tuples are ignored when computing path kernels. This model should be highly relevant to problems where a structure is large in size, and only some sub-structures are relevant for a given task; for instance, in biomedical domain, explaining a biological phenomenon may require either long sentences or multiple sentences.

To address the scalability issue, we introduce an efficient stochastic algorithm for learning the parameters of the convolution kernels. Generally speaking, exact learning of those parameters would require constructing an $N \times N$ Gram matrix of pairwise similarities between the labeled data points, which become computationally expensive for large $N$. Instead, here we propose an appropriate loss function defined over a k-NN graph,[1] an approach similar to that used for distance metric learning (Weinberger et al., 2006; Weinberger and Saul, 2009). Kernel-based locality-sensitive hashing (LSH) allows computing a k-NN graph approximately,[2] with as low as $O(N^{1.5})$ (Kulis and Grauman, 2012). (Kulis and Grauman, 2012). Since k-NN graphs are built several times during learning, hashing may not be enough for scalable learning, and may therefore necessitate the use of our stochastic-sampling on k-NN graphs in the proposed algorithm.

Our main contributions are as follows:

(i) We propose *a nonstationary extension to the conventional kernels to achieve better expressiveness* and flexibility for general NLP tasks;

(ii) We introduce *an efficient stochastic algorithm for learning the parameters* of the convolution kernel, optionally using any kernel-LSH;

(iii) We validate our approach in experiments, for the task of structured prediction described above in reference to Fig. 1, and *find a significant increase in accuracies* across multiple datasets containing 100k (unique) labeled discrete structures.

---
[1] In this work, k-NNs give better accuracy than SVMs.
[2] Known to work in computer vision, but unexplored in NLP.

Table 1: The paths of tuples are obtained by traversing the subgraphs, in Fig. 1(b) and 1(c), respectively.

| |
|---|
| **Positive labeled:** ~~(protein, example) (protein, arg0)~~ (have, arg1-of) (and, op1-of) (site, purpose-of) (bind) (bind) (arg1, enzyme) (bind) (purpose-of, site) (op1-of, and) ~~(op2, site)~~ (purpose, exchange) (arg1, small-molecule) |
| **Negative labeled:** (enyzme, arg1) (bind) (bind) (purpose-of, site) (op1-of, and) ~~(op2, site)~~ (purpose, exchange) (arg1, small-molecule) (bind) (purpose-of, site) (op1-of, and) ~~(op2, site) (arg2, small-molecule)~~ |

In addition to the above contributions, another novelty of our work is the use of kernel-LSH in NLP, although we note that (non-kernel) LSH has been used for NLP tasks such as documents retrieval, machine translation, etc (Bawa et al., 2005; Li et al., 2014; Wurzer et al., 2015; Shi and Knight, 2017).

## 2 Problem Statement

In Fig. 1, a text sentence "Furthermore, GEFs such as Sos have allosteric sites for Ras binding as well as sites for GDP/GTP exchange, and it is hard to measure GTP loading on individual Ras isoforms in cells", is parsed into an Abstract Meaning Representation (AMR) graph (Banarescu et al., 2013); we use the translation based AMR parser proposed in (Pust et al., 2015). We extract structured information about biomolecular interactions, each involving the interaction type, and two or three participants with a catalyst or domain role, from the AMR graph (Garg et al., 2016).

In the AMR graph, the blue nodes are for potential interaction types, and the green nodes are potential participants. Based on the list of interaction types, and participants, multiple candidate bio-molecular interactions are generated automatically; these are referred as the candidates for structured prediction, each one to be binary classified. As per the meaning of the source text, a candidate is manually annotated as positive (correct) or negative (incorrect), for training/test purposes.

As in the previous work by (Garg et al., 2016), for a given candidate, a corresponding subgraph is extracted from the AMR, and post-processed such that the interaction-type becomes the root node in the subgraph, and the participants become leaf nodes; see Fig. 1(b), 1(c). In the subgraphs, yellow color is for the catalyst role and green for the domain role.

Thus, in order to classify a candidate bio-molecular interaction, its corresponding subgraph can be classified as a proxy; same applies to the labeled candidates in a train set. Also, from a subgraph, a sequence can be generated by traversals between the root node and the leaf nodes.

In general, we are interested in classification of natural language structures such as sequences and graphs.

## 3 Related Work

Nonstationary kernels have been explored for modeling spatiotemporal environmental dynamics or time series relevant to health care, finance, etc., in which the challenge in learning is due to the high number of local parameters, scaling linearly with dataset size (Paciorek and Schervish, 2003; Assael et al., 2014). The number of nonstationary parameters in our NLP model is a small constant, though the cost of computing kernels is much higher compared to the kernels relevant for those domains.

Low rank approximations reduce the training/inference cost of a kernel-based classifier, but not the the cost of computing kernels (Williams and Seeger, 2001; Hoang et al., 2015). Also, caching reduces kernel cost if there are high redundancies in the substructures (Severyn and Moschitti, 2012). For reducing kernel cost in inference, *though not in learning*, approximations exist for the subselection of data (Tibshirani, 1996; Das and Kempe, 2011). For k-NN graphs, most of the approximations, including many variants of locality-sensitive hashing (LSH), have been built for explicit features space (Indyk and Motwani, 1998; Heo et al., 2012). Recently, LSH was extended for kernels (Kulis and Grauman, 2012; Raginsky and Lazebnik, 2009). There is another kernel based hashing model that employs a Support Vector Machine to obtain a random maximum margin boundary in the kernel implied feature space for generating each of the hash code bits (Joly and Buisson, 2011).

While our nonstationary modeling and learning approach is generic enough for extending any convolution kernel for an NLP task (Collins and Duffy, 2002; Moschitti, 2006; Qian and Zhou, 2012; Srivastava et al., 2013; Hovy et al., 2013; Filice et al., 2015; Tymoshenko et al., 2016), bioinformatics is one of the important problem-domains in which convolution kernels have been explored with some success (Mooney and Bunescu, 2005; Tikk et al., 2010; Airola et al., 2008; Garg et al., 2016; Rao et al., 2017). Recurrent neural networks have also been applied for extracting information on pairs of interacting proteins (Hsieh et al., 2017) though it is shown in their work, through cross corpus evaluations, that kernel methods are more robust in domain adaptation scenarios compared to the neural language models. Outside the biomedical domain, SemEval-2010 Task 8 (Hendrickx et al., 2009) is one of the popular relation extraction task (Socher et al., 2012; Li et al., 2015; Hashimoto et al., 2015; Xu et al., 2015a; Xu et al., 2015b; Santos et al., 2015; Miwa and Bansal, 2016) wherein accuracy numbers are relatively higher as the relations between entities span across short text, making the task simpler.

In neural language models (Hochreiter and Schmidhuber, 1997; Mikolov et al., 2010; Sundermeyer et al., 2012), it is challenging to remember a long sequence (Mikolov et al., 2014), as well as forgetting some parts of it (Gers et al., 1999; Zhang et al., 2015; Yu et al., 2017), with some success attained in the recent past on these problems. On the other hand, convolution (path or subsequence) kernels, implicitly, try to remember entire sequences (leads to less efficient language modeling if text is noisy), and our work allows skipping some of the tuples in sequences, i.e. forgetting.

Our idea of skipping tuples in convolution kernels had been previously explored relying on substructure mining algorithms (Suzuki et al., 2004; Suzuki and Isozaki, 2006; Severyn and Moschitti, 2013). Recently, it is proposed to learn weights of sub-structures for regression problems within Gaussian process modeling framework (Beck et al., 2015). In this paper, we show that our principled approach of nonstationary extension of a convolution kernel leads to an additional parameterization that allows learning the skipping of tuples as a special case. Further, for k-NN classifiers, we propose an efficient stochastic sampling based algorithm, along with an appropriate loss function, for scalable learning of the parameters with hundreds of thousands of data in a training set. In contrast, the previous works do not scale beyond a couple of thousand data in a train set. Moreover, as reported in (Beck et al., 2015), the introduced parameterization can lead to overfitting in their work whereas our stochastic sampling based learning ensures robustness to such issues. Also note that, our approach of skipping tuples through non-stationarity is more explicit in ignoring sub-structures irrelevant for a given task, and complementary to the standard skipping over non-matching tuples that is a common aspect in most of the existing (sparse) convolutions kernels (Zelenko et al., 2003; Mooney and Bunescu, 2005; Tkachenko and Lauw, 2015).

It is also interesting to note that most of the previous works, which explored convolution kernels for various NLP tasks, used Support Vector Machine (SVM) classifiers. The use of a kernel-SVM would have been appropriate if a convolution kernel projects natural language structures into a higher dimensional (kernel implied) features space wherein the structures from different classes were linearly separable. Unfortunately, unlike kernels operating on non-structured inputs, the existing convolution kernels (meant for structured inputs) are not guarantied to have such properties of higher dimensional mappings. This motivates us to explore k-NN like non-linear classifiers in this work as those have non-linear class boundaries even in the kernel implied feature space, in contrast to SVMs.[3]

## 4 Convolution Kernels and Locality Sensitive Hashing for k-NN Approximations

Many NLP tasks involve a classification problem where, given a set of discrete structures $\mathcal{X}$ and the associated class labels $\mathcal{Y}$, $(\mathcal{X}, \mathcal{Y}) = \{X_i, Y_i\}_{i=1}^N$, the goal is infer the correct label $Y_*$ for a test structure $X_*$. Our approach for the problem is convolution kernel (Sec. 4.1) based k-NN classifiers (Sec. 4.2).

### 4.1 Convolution Kernels

Convolution kernels belong to a class of kernels that compute similarity between discrete structures (Haussler, 1999; Collins and Duffy, 2002). In essence, convolution kernel similarity function $K(X_i, X_j)$ between two discrete structures $X_i$ and $X_j$, is defined in terms of function $k(.,.)$ that characterizes similarity between a pair of tuples or labels.

---
[3]The use of kNN models has been discouraged in high dimensional euclidean spaces, due to the difficulty of learning appropriate distance/similarity functions; however, such difficulties should not arise for structured inputs.

**Path/Subsequence Kernels:** Let $S_i$ and $S_j$ be two sequences of tuples, as in Table 1. (Mooney and Bunescu, 2005) define the kernel as:

$$K(S_i, S_j) = \sum_{\boldsymbol{i},\boldsymbol{j}:|\boldsymbol{i}|=|\boldsymbol{j}|} \prod_{k=1}^{|\boldsymbol{i}|} k(S_i(\boldsymbol{i}_k), S_j(\boldsymbol{j}_k)) \lambda^{l(\boldsymbol{i})+l(\boldsymbol{j})}.$$

Here, $k(S_i(\boldsymbol{i}_k), S_j(\boldsymbol{j}_k))$ is the similarity between the $k_{th}$ tuples in the subsequences $\boldsymbol{i}$ and $\boldsymbol{j}$, of equal length; $l(.)$ is the actual length of a subsequence in the corresponding sequence, i.e., the difference between the end index and start index (subsequences do not have to be contiguous); $\lambda \in (0, 1)$ is used to penalize the long subsequences.

**Tree Kernels:** (Zelenko et al., 2003) define the similarity between two graphs $T_i$ and $T_j$, as in Table 1(b) and 1(c) as

$$K(T_i, T_j) = k(T_i.r, T_j.r)(k(T_i.r, T_j.r) + \sum_{\boldsymbol{i},\boldsymbol{j}:l(\boldsymbol{i})=l(\boldsymbol{j})} \lambda^{l(\boldsymbol{i})}$$
$$\sum_{s=1,\cdots,l(\boldsymbol{i})} K(T_i[\boldsymbol{i}[s]], T_j[\boldsymbol{j}[s]]) \prod_{s=1,\cdots,l(\boldsymbol{i})} k(T_i[\boldsymbol{i}[s]].r, T_j[\boldsymbol{j}[s]].r)),$$

where $\boldsymbol{i}, \boldsymbol{j}$ are child subsequences under the root nodes $T_i.r, T_j.r$ and $\lambda \in (0, 1)$ as shown above; $\boldsymbol{i} = (i_1, \cdots, i_l)$ and $T_i[\boldsymbol{i}[s]]$ are subtrees rooted at the $\boldsymbol{i}[s]_{th}$ child node of $T_i.r$.

For both kernels above, dynamic programming is used for efficient computation.

### 4.2 Hashing for Constructing k-NN Graphs

A k-NN classifier finds the k-nearest neighbors of a data point per a given kernel function, and then infers its class label from the class labels of the neighbors.

In computer vision domain, a well-known technique is to approximate a kernel-k-NN graph using kernel-locality-sensitive hashing (kernel-LSH or KLSH). A smaller subset of data points, $\mathcal{A}$, of size $M$, is randomly sub-selected from the dataset $\mathcal{X}$ of size $N$, with $M$ as low as $N^{\frac{1}{2}}$. For computing a hash code of $H$ binary bits of a data point, its kernel similarity is computed w.r.t. the data points in the set $\mathcal{A}$, as an input to any kernel-LSH algorithm; a binary hashcode corresponds to a hash bucket. In the final step, the nearest neighbors of a data point are found by computing its kernel similarity w.r.t. the data points in the same hash bucket, and optionally the neighboring ones.

In the following, we brief on two KLSH approaches below. One of the approaches is proposed by (Kulis and Grauman, 2009), and the another one we develop as a *minor contribution* in this paper.

---

**Algorithm 1** Our Random kNN based Algorithm for Computing KLSH Codes.

---
**Require:** Kernel similarity vector $\boldsymbol{k}$ of size $M$, computed for a structure $S$; sub-sampling parameter, $\alpha$, for randomizing hash function; the number of hash functions, $H$. #Random subsets, fixed across structure inputs
1: **for** $j = 1 \to H$ **do**
2:     $\boldsymbol{r}^h_{j_1} \leftarrow$ randomSubset$(M, \alpha)$
3:     $\boldsymbol{r}^h_{j_2} \leftarrow$ randomSubset$(M, \alpha)$
4: **end for**
5: $\boldsymbol{c} \leftarrow \boldsymbol{0}$
6: **for** $j = 1 \to H$ **do**
7:     $\boldsymbol{k}_{j_1} \leftarrow \boldsymbol{k}(\boldsymbol{r}^h_{j_1})$
8:     $\boldsymbol{k}_{j_2} \leftarrow \boldsymbol{k}(\boldsymbol{r}^h_{j_2})$
9:     **if** $\max(\boldsymbol{k}_{j_1}) > \max(\boldsymbol{k}_{j_2})$ **then**
10:        $\boldsymbol{c}(j) \leftarrow 0$
11:     **else**
12:        $\boldsymbol{c}(j) \leftarrow 1$
13:     **end if**
14: **end for**
15: **return** $\boldsymbol{c}$

---

**KLSH by (Kulis and Grauman, 2009):** One approach to building a binary hash function is to randomly sample a linear-hyperplane in the feature space implied from the kernel functions; for $\mathcal{A}$ of size sublinear in the size of dataset $\mathcal{X}$, one can obtain only an approximation for the random sampling of linear hyperplane (Kulis and Grauman, 2012). This approach has limitations in terms of computational cost and numerical instability. The number of data points in different buckets can be highly uneven with this hash function, as per our experiments, leading to an increase in kernel computations.

**Our Proposed Random kNN based KLSH:** To build a random binary hash function $h_j(.)$, we randomly sample two subsets, $\mathcal{H}^1_j, \mathcal{H}^2_j$, from $\mathcal{A}$. Having $\mathcal{H}^1_j, \mathcal{H}^2_j$, and a given data point $X_i$, the first nearest neighbor of $X_i$ is found in both of the subsets using the convolution kernel. Depending on which of the two data points is the closest to $X_i$, a binary bit value is obtained. In our experiments, we find this hashing function as accurate as the above, yet much cheaper in kernel computations.[4] Our hashing approach is similar to the one, based on SVMs, proposed in (Joly and Buisson, 2011). See the pseudo code for our KLSH approach in Algorithm 1.

## 5 Nonstationary Convolution Kernels

We propose a generic approach to extend any convolution kernel to a non-stationary one, for higher expressibility and generalization.

**Definition 1** (Stationary kernel (Genton, 2001)). *A stationary kernel, between vectors $x_i, x_j \in \mathbb{R}^d$, is the one which is translation invariant:*

$$k(x_i, x_j) = k^S(x_i - x_j),$$

*that means, it depends only upon the lag vector between $x_i$ and $x_j$, and not the data points themselves.*

For NLP context, stationarity in convolution kernels is formalized in Theorem 1.

**Theorem 1.** *A convolution kernel $K(.,.)$, a function of the kernel $k(.,.)$, is stationary if $k(.,.)$ is stationary.*

*Proof Sketch.* Suppose we have a vocabulary set, $\{l_1, \cdots, l_p, \cdots, l_{2p}\}$, and we randomly generate a set of discrete structures $\mathcal{X} = \{X_1, \cdots, X_N\}$, using $l_1, \cdots, l_p$. For kernel $k(.,.)$, that defines similarity between a pair of labels, consider a case of stationarity, $k(l_i, l_j) = k(l_{i+p}, l_j) = k(l_i, l_{j+p}); i, j \in \{1, \cdots, p\}$, where its value is invariant w.r.t. to the translation of a label $l_i$ to $l_{i+p}$. In the structures, replacing labels $l_1, \cdots, l_p$ with $l_{p+1}, \cdots, l_{2p}$ respectively, we obtain a set of new structures $\bar{\mathcal{X}} = \{\bar{X}_1, \cdots, \bar{X}_N\}$. Using a convolution kernel $K(.,.)$, as a function of $k(.,.)$, we obtain same (kernel) Gram matrix on the set $\bar{\mathcal{S}}$ as for $\mathcal{S}$. Thus $K(.,.)$ is also invariant w.r.t. the translation of structures set $\mathcal{S}$ to $\bar{\mathcal{S}}$, hence a stationary kernel (Def. 1). □

---

[4]The number of data points across different buckets is relatively uniform with our KLSH algorithm, compared to the one in (Kulis and Grauman, 2009), since kNN models are highly data driven. This keeps the number of kernel computations low, when finding nearest neighbors in neighboring buckets. Also, unlike their algorithm, our algorithm doesn't require computing a kernel matrix between the data points in $\mathcal{A}$.

Some examples of kernels $k(.,.)$ are:

$k(i,j) = \mathbb{I}(i = j)$, $k(i,j) = \exp(-||w_i - w_j||_2^2)$,
$k(i,j) = \exp(w_i^T w_j - 1)$.

Herein, $i, j$ are words (node/edge labels), and $w_i, w_j$ are word vectors respectively; $\mathbb{I}(.)$ is an indicator function. The first two kernels are stationary. Whereas the third one is nonstationary mathematically (yet not expressive enough, as such), though it relies only on the cosine similarity between the word vectors, not the word-vectors or words themselves. Our approach is generic enough for not only extending a stationary convolution kernel, but a nonstationary convolution kernel as well, accounting for the cases as the latter one, *for a higher flexibility and generalization.*

**Theorem 2.** *Any convolution kernel $K(.,.)$, can be extended to a valid nonstationary convolution kernel $K^{NS}(.,.)$, by extending the stationary/nonstationary kernel function $k(.,.)$ to $k^{NS}(.,.)$ as in (1), with a deterministic function $\sigma(.)$ of any form.*

$$k^{NS}(i, j) = \sigma(i) k(i, j) \sigma(j) \quad (1)$$

*Proof Sketch.* The extended kernel function $k^{NS}(i, j)$ is a valid kernel (Rasmussen, 2006, p. 95), and therefore, $K^{NS}(.,.)$, as a function of $k^{NS}(.,.)$, is also a valid convolution kernel. For establishing the nonstationarity property, following the proof of Theorem 1, if using $K^{NS}(.,.)$, we obtain a (kernel) Gram matrix on the set $\bar{\mathcal{S}}$ that is different from the set $\mathcal{S}$ because $\sigma(l_i) \neq \sigma(l_{i+p}) \; \forall i \in \{1, \cdots, p\}$ for an arbitrary selection of $\sigma(.)$. Therefore $K^{NS}(.,.)$ is not invariant w.r.t. the translation of set $\mathcal{S}$ to $\bar{\mathcal{S}}$, hence a nonstationary kernel (Def. 1). □

In natural language modeling problems, since $k(.,.)$ operates on labels (or a tuple of labels), $\sigma(.)$ can be thought of as 1-D embedding of the vocabulary for labels. For $\sigma(.) \in \mathbb{R}_{\geq 0}$, it is like the strength of a label in the context of computing the convolution kernel similarity. *For instance, if $\sigma(i) = 0$, it means that the label $i$ should be completely ignored when computing a convolution kernel similarity of a discrete structure (tree, path, etc.), that contains the (node or edge) label $i$, w.r.t. another discrete structure, or itself.* Thus, we see that these additional parameters, introduced with the nonstationary model, allow convolution kernels to be expressive enough to decide if some substructures in a discrete structure should be ignored. One can also define $\sigma(.)$ on the vector space representation of the labels (word vectors), like $\sigma(\boldsymbol{w}_i)$ instead of $\sigma(i)$.

## 5.1 Nonstationarity for Skipping Sub-Structures

Suppose we are interested in computing the convolution kernel similarity between the paths of tuples — as in Table 1 or between the two trees in Table 1(b), 1(c). In both cases, the basic kernel function, $k(.,.)$ operates on a pair of tuples. The nonstationary kernel on tuples can be defined as:

$$k^{NS}((e_i, n_i),(e_j, n_j)) = \sigma_{e_i} k_e(e_i, e_j) \sigma_{e_j} k_n(n_i, n_j).$$

The similarity between two tuples $(e_i, n_i)$ and $(e_j, n_j)$ is defined as the product of kernels on the edge labels $e_i, e_j$ and the node labels $n_i, n_j$. In our experiments, we consider the following functions:

$$k_e(e_i, e_j) = \mathbb{I}(e_i = e_j), \ k_n(n_i, n_j) = \mathbb{I}(n_i = n_j),$$
$$k_n(n_i, n_j) = \exp(\boldsymbol{w}_{n_i}^T \boldsymbol{w}_{n_j} - 1)((\boldsymbol{w}_{n_i}^T \boldsymbol{w}_{n_j} - \gamma)/(1-\gamma))_+.$$

Herein, $(.)_+$ denotes the positive part; $\gamma \in (-1, 1)$ is a sparsity parameter.

We consider the nonstationary parameters to be binary valued, i.e., $\sigma \in \{0, 1\}$ (we keep binary values only, as a measure for robustness to over-fitting, as well as for keeping the optimization simple).

In the expression above, non-stationarity parameters are local to edge labels (referred as semantic labels in this paper), which come from a language representation (predefined, and small in vocabulary), such as Abstract Meaning Representations (Banarescu et al., 2012), Stanford Dependencies (Chen and Manning, 2014), etc. Defining the $\sigma$ parameters local w.r.t. the edge labels (semantic labels), corresponds to attention on the semantics in the context of convolution kernel computation. For instance, as shown in Table 1, we find that it is more optimal to skip tuples, containing either of the semantic labels *arg2, arg0, op2, example* when computing convolution kernel similarity, as $\sigma = 0$ for all these labels in our learned model for the task .[5] Thus, our proposed nonstationary approach allows decisions on what to attend in a linguistic representation of a natural language, and what not to attend. This is very intuitive, as it is close to how a human mind processes natural language.

---
[5]arg0, arg2 are not ignored if duplicate data points are kept.

Note that, even in the traditional convolution kernels, matching substructures are found while skipping over non-matching elements. Our approach of skipping tuples, as proposed above, allows skipping tuples even if those tuples are matching when computing similarity between two structures. This aspect of skipping tuples is more explicit in ignoring substructures irrelevant for a given task, unlike the standard skipping over non-matching tuples; the latter is complementary to ours.

## 6 Stochastic Learning of Kernel Parameters

In this section we discuss the learning of kernel parameters for classification with k-Nearest Neighbor models.

### 6.1 k-NN based Loss Function

In the above, we introduced a non-stationary extension of kernel similarity by introducing a task-dependent parameterization of the kernel function. In this section we describe our approach for efficiently learning those parameters in the context of a binary classification problem, especially suitable for k-NN classifiers. This setup is highly relevant for various structured inference problems in natural language modeling, where a hypothesized structured inference from a sentence is classified as correct/incorrect. Note that the loss-function proposed below can be easily extended to multi-class scenarios.

**Loss Function 1.** *Consider $N$ number of training examples—say, sequences of tuples $S_1, \cdots, S_N$, as we show in Table 1, with corresponding binary labels $y_1, \cdots, y_N \in \{0, 1\}$. A convolution kernel $K(.,.; \boldsymbol{\sigma})$ with parameters $\boldsymbol{\sigma}$ can be learned by minimizing the loss function,*

$$\mathcal{L} = \sum_{i=1}^{N} K(S_i, S_{1nn(i)}; \boldsymbol{\sigma}) \mathbb{I}(y_i \neq y_{1nn(i)})$$
$$- \sum_{i=1}^{N} K(S_i, S_{1nn(i)}; \boldsymbol{\sigma}) \mathbb{I}(y_i = y_{1nn(i)} = 1).$$

**Algorithm 2** Stochastic-subsampling based minimization of the loss, defined on k-NN, for learning a convolution kernel

**Require:** Training data set $S = \{S_1, \cdots, S_N\}$; the number of samples to select randomly, $\beta$; the number of trials for optimizing a parameter value, $\alpha$; the convolution kernel function $K(.,.;\sigma)$; the semantic-labels $L = \{l_1, \cdots, l_p\}$; $k$ for the global k-NN graph.

```
#indices of data points, c_i, containing a semantic label, l_i
```
1: $c_1, \cdots, c_p \leftarrow getDataIndices(S, L)$
2: $\sigma = \{\sigma_1, \cdots, \sigma_p = 1\}$ `#parameters, sorted per the frequency of the labels in descending order`
   `#global k-NN graph for intelligent sampling`
3: $G \leftarrow computeNNGraph(S, k, K, \sigma)$
   `#optimize each t_th parameter, iteratively`
4: **for** $t = 1$ to $p$ **do**
5:   **for** $j = 1$ to $\alpha$ **do**
6:     $r \leftarrow randomSubset(c_t, \beta)$ `#β random samples`
7:     $n \leftarrow getNN_{\rightarrow}(G, r, k)$ `#neighbors of r`
8:     $\bar{n} \leftarrow getNN_{\leftarrow}(G, r, k)$ `#r as neighbors`
9:     $nn \leftarrow getNN_{\rightarrow}(G, n \cup \bar{n}, 1)$ `#neighbors of n ∪ n̄`
10:     $a \leftarrow r \cup n \cup \bar{n} \cup nn$ `#a for loss estimates`
11:     **for** $v = \{0, 1\}$ **do**
12:       $\sigma_t = v$ `#value v for the t_th parameter`
          `#noisy loss estimate for K(.,.;σ)`
13:       $G_{tj}^v \leftarrow computeNNGraph(S_a, 1, K, \sigma)$ `#1-NN graph on the subset a for loss estimate`
14:       $\mathcal{L}_{tj}^v \leftarrow computeLoss(G_{tj}^v)$ `#loss on G_{tj}^v`
15:     **end for**
16:   **end for**
17:   $\sigma_t \leftarrow \arg\min_v \sum_{j=1}^{\alpha} \mathcal{L}_{tj}^v$
18: **end for**
19: **return** $\sigma = \{\sigma_1, \cdots, \sigma_p\}$ `#learned σ, in K(.,.;σ)`

For a sequence $S_i$, we correspondingly have the first nearest neighbor of $S_i$ within the training data set itself, i.e., $S_{1nn(i)}$, with a binary label $y_{1nn(i)}$; the function $1nn(.)$ comes from kernel-based 1-NN graph construction, optionally using kernel-LSH.

If the binary label of $S_i$, $y_i$, does not match with the binary label of $S_{1nn(i)}$, $y_{1nn(i)}$, then there is loss, equal to the convolution kernel similarity $K(S_i, S_{1nn(i)}; \sigma)$. In the loss function, we also encourage higher kernel similarity between a sequence and its first nearest neighbor if both are of positive (binary) labels by using the negative loss term, i.e., reward. The latter should make the kernel function robust w.r.t. noise in data, acting more like a regularization term.

The loss function is also extensible for $k > 1$, though we argue that it should be sufficient to keep $k = 1$ during the learning,[6] as the loss function is computed using a small subset of training dataset with our stochastic approximation-based algorithm (discussed next).[7]

Note that the loss function is similar to the ones used for distance metric learning problems (Weinberger and Saul, 2009).

### 6.2 Stochastic Subsampling Algorithm

The loss function defined above involves calculating $O(N^2)$ pairwise similarities between $N$ data points, which is computationally prohibitive for large $N$. Instead, we would like to approximate the loss function by using a small subset of the training data. The naive strategy of random sampling would correspond to taking a random subset of nodes in the k-NN graph, and then considering the neighborhood edges only between those nodes. Unfortunately, uniform random sampling of nodes (data points) would be a crude approximation for the loss function as it would significantly alter (or even completely destroy) local neighborhood structure for each selected data point. Instead, here we propose to include the

---
[6] $k = 1$ in the loss function makes LSH efficient, too.

[7] A bagged 1-NN classifier is optimal, as a k-NN model with an optimal value of $k$ (Biau et al., 2010).

Table 2: F1 score (precision, recall), on the positive class. Our non-stationary path kernel (NPK), trained on a subset of the "All" dataset with our hashing function. For inference, $k = 8$ ($k = 4$ for PubMed45 since it is a small data set) in k-NN based classifiers, with hashing used in both approaches "Path Kernel" (PK) and "Nonstationary Path Kernel" (NPK). We perform five inference trials, for the cases of 80%, 50%, 30% data sub-selection.

(a) Our hashing function

| Data sets | PK | NPK |
|---|---|---|
| BioNLP (100%) | 42.4 (43.1, 41.8) | **47.2** (**53.7**, 42.2) |
| BioNLP (80%) | 41.0 ± 0.5 (40.7, 41.4) | **43.9 ± 1.3** (**47.7**, 40.7) |
| PubMed45 (100%) | 38.7 (37.8, 39.6) | **44.7** (**45.8, 43.7**) |
| PubMed45 (80%) | 36.3 ± 0.9 (35.7, 36.8) | **41.8 ± 1.1** (**42.4, 41.3**) |
| All (100%) | 55.0 (52.5, 57.8) | **58.4** (**55.0, 62.2**) |
| All (50%) | 50.9 ± 0.6 (49.5, 52.4) | **54.1 ± 0.7** (50.0, **59.0**) |
| All (30%) | 48.6 ± 1.2 (46.7, 50.6) | **51.5 ± 0.9** (48.4, **55.1**) |

(b) Hashing function of Kulis & Grauman.

| Data sets | PK | NPK |
|---|---|---|
| BioNLP (100%) | 44.8 (45.7, 44.0) | **46.8** (**52.7**, 42.1) |
| BioNLP (80%) | 42.0 ± 1.0 (41.7, 42.2) | **44.6 ± 1.5** (**49.3**, 40.9) |
| PubMed45 (100%) | 38.6 (39.1, 38.1) | **43.5** (**44.9, 42.2**) |
| PubMed45 (80%) | 35.5 ± 0.8 (35.6, 35.4) | **41.7 ± 1.1** (**42.0, 41.4**) |
| All (100%) | 54.2 (52.6, 55.9) | **57.9** (**54.6, 61.7**) |
| All (50%) | 52.2 ± 0.4 (50.6, 53.9) | **54.3 ± 0.6** (50.9, **58.2**) |
| All (30%) | 50.3 ± 0.6 (48.7, 52.1) | 51.6 ± 0.8 (48.2, 55.6) |

neighboring nodes of a randomly selected node in the subset of data used for computing a loss estimate. Thus, the problem is how to select the local neighborhoods, which depend upon the kernel function itself that we are optimizing. We accomplish this as described below.

First, before optimizing the parameters, we compute the nearest neighborhood graph for all the training data points with a locality-sensitive hashing technique and by using the initialized convolution kernel function itself; we refer to this as a global k-NN graph, used for building the local neighborhoods approximately.

Now we sample a small subset of data from the training set in a purely random fashion, as the inducing points. Then we look for the training data which are neighbors of these random selections, or have these random selections themselves as their neighbors in the k-NN graph; optionally, we also try to include the first nearest neighbors of the neighbors. Putting all these data points together into a subset, including the original subset of random selections, we obtain our final subset of the training dataset—what we referred to as randomly sampled local neighborhoods. We compute a 1-NN graph on this subset of the dataset itself to compute a noisy estimate of the loss function. As we see, the neighbors (in both directions) of an inducing point (randomly selected) establish the local neighborhood around it. Even though, in the loss function, we are interested only in a small neighborhood around a point, i.e., defined from the first nearest neighbor, we keep $k$ reasonably high in the above global k-NN graph and correspondingly obtain $k$ neighbors of the inducing point from both directions. This is because the higher value of $k$ ($k = 4$ in our experiments, though robust w.r.t. small changes) should make the approach robust to changes in the neighborhood during the learning (as parameter values change), and also increase the chances for the neighbors of an inducing point to be neighbors of each other (for less noise in the loss function estimation).

### 6.2.1 Pseudocode for the Algorithm

The pseudocode for optimizing the semantic-label- based binary parameters is presented in Algorithm 2. Since a parameter $\sigma_t$ is defined locally w.r.t. a semantic label $l_t$, we obtain the indices of data points ($c_t$), in the training set $S$, that contain the semantic label; these data should be immediately relevant for the optimization of the parameter as the superset of the inducing points (see line 1 in Algorithm 2). All the binary parameters,

Table 3: F1 score on the positive class. Same settings as for Table 2(a).

(a) NPK with lower sampling rates ($\beta$)

| Data sets | $\beta = 550$ | $\beta = 100$ | $\beta = 50$ | $\beta = 25$ |
|---|---|---|---|---|
| BioNLP | **47.2** (53.7, 42.2) | **47.3** (52.5, 43.0) | 47.1 (52.4, 42.8) | **48.4** (50.9, 46.0) |
| PubMed45 | **44.7** (45.8, 43.7) | 42.7 (45.2, 40.4) | 43.0 (42.2, 43.8) | 41.9 (41.8, 42.0) |

(b) TK ($k = 1$)

| TK | NTK |
|---|---|
| 38.5 (37.6, 39.4) | **41.8** (42.0, 41.5) |
| 33.4 (33.0, 33.8) | **37.8** (36.6, 39.0) |

(c) With word2vec

| Data sets | PK | NPK |
|---|---|---|
| BioNLP | 40.8 (41.5, 40.2) | **43.7** (44.9, 42.5) |
| PubMed45 | 35.0 (34.0, 36.1) | **36.5** (37.7, 35.3) |

(d) Nonunique

| PK | NPK |
|---|---|
| 60.2 (62.2, 58.3) | **62.4** (**65.2**, 59.8) |
| 36.6 (44.1, 31.3) | **40.5** (45.9, **36.2**) |

(e) Individual Training

| PK | NPK |
|---|---|
| 41.8 (43.0, 40.7) | **44.5** (44.0, **45.0**) |
| 38.1 (39.0, 37.3) | **42.8** (**42.9**, **42.7**) |

$\sigma = \{\sigma_1, \cdots, \sigma_p\}$, are initialized to value one, which corresponds to equal attention to all types of semantics in the beginning, i.e., the baseline convolution kernel (line 2). During the optimization, as we would learn zero values for some of the parameters, the kernel function would drift away from the baseline kernel and would become more non-stationary, accounting for only a subset of the semantic labels in its computations.

Next, after the initialization of the binary parameters, we compute the global k-NN graph on the training data set $S$, using the kernel $K(.,.;\sigma)$ with the parameters $\sigma$ (line 3).

For learning these binary parameters, $\sigma$, we find that a simple iterative procedure does a good job (line 4), though the algorithm should be easily extensible for more sophisticated techniques such as MCMC sampling. For optimizing a $t_{th}$ parameter, $\sigma_t$, we perform $\alpha$ number of trials (line 5).

In each trial, a subset of the training set, with indices $a$, is obtained using our sampling technique (lines 6-10), starting with $\beta$ number of randomly selected inducing points. The next step within the trial is to compute the loss function (line 14) for each possible value of $\sigma_t$ (i.e., $\{0, 1\}$) by computing a 1-NN graph on the subset of samples $S_a$ (line 13), and then obtaining the value of $\sigma_t$ with a minimum sum of the noisy loss estimates (line 17).

In the algorithm, the global k-NN graph (line 3), and the 1-NN graphs (line 13) can be computed with the kernel-LSH, as we do in our experiments; the same applies to inference after learning the parameters.

Next, we discuss our empirical results.

## 7 Empirical Evaluation

For the evaluation of our nonstationary convolution kernels, we consider the following task that is also described in Sec. 2.

**Structure Prediction from Semantic Graphs:** In reference to Fig. 1, each sentence is parsed into an Abstract Meaning Representation (AMR) graph (Pust et al., 2015), and from the list of entity nodes and interaction type nodes in the graph, a set of candidate biomolecular interactions are generated—each involving up to three entities with different roles, and an interaction type. Each candidate structure (a biomolecular interaction in this case), as a hypothesized inference, is classified as positive/negative, either using the corresponding subgraph as features (as in Fig. 1(b), 1(c)) with tree kernels, or a subgraph can be post-processed into a path (see Table 1) if using path kernels.

**Data Sets:** We use three datasets, including two public ones, PubMed45-ERN (Garg et al., 2016) and BioNLP (2009, 11,13) (Kim et al., 2009; Kim et al., 2011; Nédellec et al., 2013), and an unpublished one; we refer to PubMed45-ERN simply as PubMed45 in this paper though PubMed45 refers to a subset of PubMed45-ERN in (Garg et al., 2016). Putting all the datasets together, referred as "All" dataset, we have approximately 100k unique data points (20k and 30k from the PubMed45 and

Table 4: F1 score on the positive class. Same settings as for Table 2(a).

(a) $\beta = 25$, multiple runs of the algorithm for NPK

| Data sets | PK | $iter = 1$ | $iter = 3$ | $iter = 6$ | $iter = 9$ |
|---|---|---|---|---|---|
| BioNLP | 42.4 (43.1, 41.8) | 46.4 (51.7, 42.0) | 44.0 (49.8, 39.4) | 47.2 (53.7, 42.2) | 47.4 (51.0, 44.2) |
| PubMed45 | 38.7 (37.8, 39.6) | 41.9 (42.7, 41.1) | 41.9 (44.7, 39.4) | 44.7 (45.8, 43.7) | 41.1 (41.6, 40.7) |

(b) Pr- Pure Random.

| Data sets | PK | $\beta = 100$ | $\beta = 1000$ (PR) | $\beta = 550$ | $\beta = 5000$ (PR) |
|---|---|---|---|---|---|
| BioNLP | 42.4 (43.1, 41.8) | **47.3** (52.5, 43.0) | 42.4 (43.1, 41.8) | **47.2** (53.7, 42.2) | 43.7 (48.3, 39.9) |
| PubMed45 | 38.7 (37.8, 39.6) | **42.7** (45.2, 40.4) | 40.1 (40.1, 40.0) | **44.7** (45.8, 43.7) | 41.8 (44.7, 39.3) |

BioNLP datasets respectively), with uniqueness defined on the path tuples to avoid artificial redundancies in the evaluation, *although this decreases the overall accuracy numbers as canonical examples are counted only once*. We also show results for non-unique paths, in Sec. 7.1.2, which can be compared to other published results on the datasets.

**Settings for the Learning of the Proposed Model Parameters:** In our evaluation setup, we randomly divide the full dataset into two halves, and use the first half (with approximately N=50k samples) to learn the proposed nonstationary convolution kernel with our proposed Algorithm 2; we set $\beta = 550$, $\alpha = 10$, and learn $\sigma$ for the top 50% frequent semantic (edge) labels, i.e., 61 binary parameters. Whenever computing k-NN graphs, we use kernel-LSH (H=14), using our proposed hashing function in Sec. 4. [8]

For building a k-NN graph with hashing in the learning algorithm, we keep M (defined in Sec. 4.2) approximately equal to the square root of the number of nodes in k-NN; for the inference purposes, we keep M=100 fixed across all the datasets.

In the convolution kernels, $\lambda = 0.8$, which is set from preliminary tunings, as well as the use of the learning algorithm itself. For $\beta = 550$, i.e., our primary configuration on the sample size when learning the model on all the data sets together, it takes approximately 12 hours on a 16-core machine to op-timize the parameters (a kernel matrix computing is parallelized); we also try $\beta = 25$, taking an hour for learning the model.

### 7.1 Evaluation Results

The convolution path kernel with the 61 binary parameters, referred to as Nonstationary Path Kernel (**NPK**) and optimized on the training subset from the All dataset, is evaluated w.r.t. the standard (conventional) path kernel (**PK**), across the individual public datasets PubMed45 and BioNLP, as well as the All dataset. For the evaluation of the two kernels, using k-NN classifier with hashing, we split a given data set randomly into two equal parts, with the first half for building a k-NN classifier that is tested on the second half. We have multiple inference evaluations with varying percentages of data sub-sampled from each of the two parts (5 trials).

The results are presented in Table 2(a), in terms of classification accuracy (F1-score, precision, recall) for the positive class. We see significant improvement in accuracy numbers over the baseline standard path kernel, with the improvements even more significant for the two public datasets. We obtained this improvement just with the addition of the 61 binary parameters to the baseline kernel, corresponding to our NPK. It is even more interesting to note that only 20 of the 61 parameters have zero values learned—thereby solely contributing to the improvements. Since the model is trained on all the datasets put together, it is further interesting to note that the learned model performs well for the individual pub-lic datasets, despite distribution changes across data sets in terms of positive ratios as well as language itself.

---
[8] H=14 is simply from the observation that it should be a little less than the $\log_2(.)$ of the data set size so that a constant number of data points land in a hash bucket.

Table 5: Loss function convergence.

| Models | $\beta = 200$ (25) | $\beta = 500$ (25) | $\beta = 1000$ (25) | All data |
|---|---|---|---|---|
| PK | 7.7e-3±4.9e-3 | 8.0e-3±3.2e-3 | 6.2e-3±1.5e-3 | 4.5e-3 |
| **NPK** | 3.9e-3±6.5e-3 | 2.3e-3±4.4e-3 | 2.0e-3±3.2e-3 | 2e-3 |

Further, we verify if the same model parameters, learned with our hashing function (see Sec. 4), generalize to the hashing function of (Kulis and Grauman, 2012) for inference purposes. In Table 2(b), for the same inference scenarios we show the classification accuracies with their hashing function. While there is not much difference between accuracy numbers across the two hashing functions for a given path kernel, we do see that NPK outperforms PK, even with this hashing function.

### 7.1.1 Varying Parameter & Kernel Settings

**Lower Sampling Rates:** Keeping the same setup as that used for producing the results in Table 2(a), but lowering the parameter $\beta$, in Algorithm 2,[9] we obtain results shown in Table 3(a). The accuracy numbers decrease marginally with lower values of $\beta$, down to $\beta = 25$, that correspond to smaller subsets of data during the learning. This leads to more noisy estimates of the loss function, but much less computational cost. It should be noted that all of these models (our path kernels with the binary parameters), learned with high stochasticity, still outperform the standard path kernel (in Table 2(a)) significantly.

**Tree Kernels:** Using the same setup for learning (a single model trained on all the data sets) as well as inference, we also evaluate our approach for tree kernels, with the results presented in Table 3(b). Our non-stationary tree kernels (**NTK**) outperform the standard tree kernels (**TK**), though the overall numbers are not as good as with the path kernels.[10]

Also, we find it interesting that $k = 1$ is optimal in the inference for tree kernels, possibly because the dimension of the features space implied from tree kernels is very high.

**Use of Word Vectors in Path Kernels:** Our approach is generic enough for applicability to convolution kernels using word vectors. We compare NPK approach w.r.t. PK,[11] both using word vectors, and find that our nonstationary model helps in this scenario as well; see Table 3(c).[12]

### 7.1.2 Non-Unique Path Tuples

In Table 3(d), we present experimental results where duplicate path tuples are kept in the datasets ($\beta = 130$, i.e., 0.2% of training data). For the BioNLP data set, we use the BioNLP-2013-development subset for testing, and the rest for building the k-NN classifiers. For the PubMed dataset, we use six papers for testing (approx. 10% data points), and the rest for building the classifier. For both datasets, NPK give better accuracy than PK. These numbers, in Table 3(d), can also be compared to the previously published evaluation numbers on the datasets.

### 7.1.3 Analysis of the Learning Algorithm

**Analysis of the Loss Function:** In Table 5 we present the loss (rather than the F1 score) of the standard path kernel model, as well as our nonstationary path kernel model ($\beta = 100$). For computing the loss of a given (already trained) model, we use the same sampling technique that we proposed for learning, with $\beta = 200, 500, 1000$ corresponding to the different columns in the table, and also compute the loss on all the training data (with no sampling). The numbers in the last column "All data" validate that the loss function value is reduced with our nonstationary model parameters compared to the standard path kernel, from $4.5e-3$ to $2e-3$. The loss function variation w.r.t. the sampling rate $\beta$ establishes on empirical convergence of the approximate loss, computed with our sampling technique, to the loss computed on all the data.

---

[9]For $\beta \leq 50, \beta = 100$, we use H=8, H=10 respectively.

[10]One possible reason for the overall low numbers with tree kernels is this: In Table 1(b), if we exchange the colors of the entity nodes, and so the respective roles, the hypothesized interaction changes, becomes invalid from valid, while the tree remains very similar; the path would change significantly, as obtained from tree-traversal as per the node colors.

[11]Sparsity parameters, $\gamma$, is tuned to value 0.6.

[12]The overall accuracy numbers are less when using word vectors, partially because there is no domain mismatch in terms of vocabulary when splitting the set of unique path tuples randomly into training-test sets (in such scenarios, there is not much difference in accuracy, whether we use word vectors or not, be it our path kernels or the baseline path kernels).

Table 6: F1 score, on the positive class w.r.t. other classifiers. Same settings as for Table 2(a).

| Data sets | PK-k-NN | **NPK-kNN-H** | PK-SVM | LSTM | Conv. LSTM |
|---|---|---|---|---|---|
| BioNLP | **47.3** (54.6, **41.7**) | **47.2** (53.7, **42.2**) | 38.3 (61.5, 27.8) | 46.0 (65.7, 35.4) | 44.1 (63.8, 33.7) |
| PubMed45 | **45.3** (51.5, **40.4**) | **44.7** (45.8, **43.7**) | 40.2 (53.0, 32.5) | 34.3 (47.8, 26.7) | 40.9 (47.0, 36.2) |

**Multiple Runs of the Algorithm for Learning:** We also try the experiment of running the algorithm multiple times wherein, after each iteration, the global k-NN graph is recomputed as per the change in the kernel function ($\beta = 25$); the models learned across different iterations are evaluated as presented in Table 4(a). While all the models, learned across different iterations, have a much better accuracy compared to the standard path kernel (baseline), the accuracy doesn't seem to always increase with more iterations, partly due to the high noise from the low sampling rate. This should mean that the learning is not highly sensitive to the fine grained structure of the global k-NN graph, while the value of $\beta$ plays more significant role, as we get much better accuracy for $\beta = 550$ above, even with a single iteration, than the model obtained from 10 iterations of learning with $\beta = 25$.

**Experiments with Pure Random Sampling:** While keeping the same nonstationary model, a possible baseline approach for the learning could be sampling the data in a pure random fashion rather than using our sampling technique; see Table 4(b). For this baseline, we sample only the $\beta$ inducing points in Algorithm 2, and not the neighbors (suffix "(PR)"). Clearly, the strategy of just pure random sampling doesn't work well.

### 7.1.4 Individual Models Learned on Datasets

For a further evaluation, we also train the model individually for each of the two public datasets, PubMed45 and BioNLP. The experimental settings are the same, using 50% random selections for training, and the rest for inference. The value for parameter $\beta$ is 3% of the training subset size (i.e., approximately, $\beta = 300$, $\beta = 450$ for the PubMed45 and BioNLP training subsets respectively); H=10 as the training data set size is smaller than above, while using our hashing function in both learning and inference tasks. The results are presented in Table 3(e). Although our NPK outperforms the standard path kernel more significantly for the PubMed45 dataset, the improvements are less than what we achieved when training the model on all the datasets together.

### 7.1.5 Comparison with Other Classifiers

Finally, we compare our proposed nonstationary path kernel-based k-NN classifier using the kernel-LSH (the same classifier as "NPK" in Table 2(a)), denoted as "NPK-kNN-H" here, w.r.t. other kernel-based classifiers, as well as neural network baselines; see Table 6. Other kernel-based classifiers, all using the standard path kernels, include k-NN without using hashing ("PK-k-NN", k=4), and SVM ("PK-SVM", C=1). All these classifiers are expensive to compute as they require full Gram matrices, with $O(N^2)$ number of kernel computations. The hashing-based version of "PK-k-NN" corresponds to "PK" in Table 2(a). Clearly, there is a significant decrease in accuracy for standard path kernels when using hashing. In Table 6 [13] we demonstrate that our nonstationary path kernel-based k-NN model ("NPK-kNN-H"), despite the hashing-based approximations, performs as good as the PK-k-NN, and both methods outperform all the other classifiers, including the neural network models LSTM and convolution-LSTM (label frequency-based 32-bit embeddings are used in the LSTM models, with "ReLU" units). The advantage of "NPK-kNN-H" over the "PK-k-NN" is the reduction in kernel computations ($O(N^{1.5})$ from $O(N^2)$),

---

[13] For k-NN, accuracy variation w.r.t. $k$ is minimal, so we do a preliminary tuning using a much smaller subset, whereas for other methods, we either do validation based tuning, or choose the best accuracy number from the parameters space.

due to hashing.

## 8 Conclusions

In this paper we proposed a novel nonstationary extension of convolution kernels by introducing a data-driven parameterization of the kernel similarity function. The extended kernels have better flexibility and expressibility of language representations, compared to conventional convolution kernels used in natural language tasks. We validated the proposed approach in a set of experiments for a structured prediction task, and observed significant improvement of accuracy numbers over the state-of-the-art methods across several data sets. The learned models are highly interpretable, as the zero values of the parameters correspond to a list of semantic labels and corresponding substructures that are ignored during kernel computation over the semantic graph. We also proposed a tractable learning method based on a stochastic-sampling algorithm, and demonstrated that keeping the sampling rate low has only a moderate adverse impact on accuracy, while yielding significant gains in computational efficiency. In our future work, within the framework of nonstationary modeling, we shall explore other kinds of parameters applicable to language modeling and the relevant sampling techniques.

## 9 Acknowledgements

This work was sponsored by the DARPA Big Mechanism program (W911NF-14-1-0364). It is our pleasure to acknowledge discussions with Guillermo Cecchi, Elif Eyigoz, Shuyang Gao, Palash Goyal, David Kale, Kevin Knight, Daniel Marcu, Daniel Moyer, Michael Pust, Anant Raj, Irina Rish, Kyle Reing. We are also grateful to the anonymous reviewers for their valuable feedback.